\documentclass[sigconf]{acmart}
\usepackage{hyperref}
\usepackage{hyperxmp}
\usepackage{colortbl}
\usepackage{booktabs} 
\usepackage{algorithm}
\usepackage{amsmath}
\usepackage{graphicx}
\usepackage{subcaption}
\usepackage{amsfonts,bm}
\usepackage{multirow}
\usepackage{color}
\usepackage{braket}
\usepackage{caption}
\usepackage{mathtools}
\usepackage{enumerate}
\usepackage{enumitem}
\usepackage{empheq}
\usepackage{calc}
\usepackage{dsfont}
\usepackage{algpseudocode}

\AtBeginDocument{%
  \providecommand\BibTeX{{%
    \normalfont B\kern-0.5em{\scshape i\kern-0.25em b}\kern-0.8em\TeX}}}

\copyrightyear{2026}
\acmYear{2026}
\setcopyright{cc}
\setcctype{by}
\acmConference[CIKM '26]{Proceedings of the 35th ACM International Conference on Information and Knowledge Management}{November 07--11, 2026}{Rome, Italy}
\acmBooktitle{Proceedings of the 35th ACM International Conference on Information and Knowledge Management (CIKM '26), November 07--11, 2026, Rome, Italy}
\acmDOI{10.1145/3799682.3840920}
\acmISBN{979-8-4007-2539-5/2026/11}

\begin{document}

\title[Quantifying Event Impacts on Time Series via Multiscale Contrastive Learning]
{Quantifying Event Impacts on Time Series\\via Multiscale Contrastive Learning}

\author{Yiming Sun}
\affiliation{%
  \institution{Rutgers University}
  \city{New Brunswick}
  \state{NJ}
  \country{USA}}
\email{yiming.sun99@rutgers.edu}

\author{Shengyu Chen}
\affiliation{%
  \institution{NEC Laboratories America}
  \city{Princeton}
  \state{NJ}
  \country{USA}}
\email{shchen@nec-labs.com}

\author{Zhengzhang Chen}
\affiliation{%
  \institution{NEC Laboratories America}
  \city{Princeton}
  \state{NJ}
  \country{USA}}
\email{zchen@nec-labs.com}

\author{Haoyu Wang}
\affiliation{%
  \institution{NEC Laboratories America}
  \city{Princeton}
  \state{NJ}
  \country{USA}}
\email{haoyu@nec-labs.com}

\author{Xiaowei Jia}
\affiliation{%
  \institution{Rutgers University}
  \city{New Brunswick}
  \state{NJ}
  \country{USA}}
\email{xj159@cs.rutgers.edu}

\author{Haifeng Chen}
\affiliation{%
  \institution{NEC Laboratories America}
  \city{Princeton}
  \state{NJ}
  \country{USA}}
\email{haifeng@nec-labs.com}

\renewcommand{\shortauthors}{Sun et al.}

\begin{abstract}
Shocks that spread through the web, such as cybersecurity breach disclosures, can abruptly disrupt financial time series and cause substantial abnormal losses. While these events are disclosed as discrete records through news reports, regulatory filings, or public databases, their consequences unfold through continuous market dynamics. This creates an event-conditioned impact prediction problem: given pre-event market history and limited event metadata, the goal is to estimate short-term post-disclosure abnormal loss rather than reconstruct the full post-event trajectory. However, most time-series forecasting models focus on endogenous regularities such as trend, seasonality, and autocorrelation, and thus struggle with rare and heterogeneous external events. The challenge is further amplified by sparse high-impact events and background market noise. We introduce EventTime, a multi-resolution framework that combines long-horizon market context, short-horizon pre-event dynamics, and event metadata. It incorporates an event fusion module that couples temporal representations with event attributes to identify relevant recent market patterns. To mitigate sparse supervision, EventTime further introduces a dynamic contrastive objective that constructs event- and time-series-aware positive and negative pairs during training. We also construct SECURE, a real-world dataset aligning cybersecurity incidents with stock-market time series and structured and LLM-derived semantic features. Experiments show that EventTime consistently outperforms state-of-the-art time-series and event-aware baselines in estimating post-event financial losses. Further analyses demonstrate more event-sensitive representations, greater robustness to incomplete metadata, and more interpretable estimates of short-term market impact following cybersecurity disclosures.
\end{abstract}

%
%




%
\maketitle

\section{Introduction}


Cybersecurity incidents have become a prominent class of web-scale external shocks. 
They are often disclosed and disseminated through news portals, regulatory filings, public breach databases, and social media, where a single report can rapidly reshape investor perception of a firm. 
Unlike endogenous market fluctuations that evolve gradually through trends, cycles, and short-term autocorrelation, such disclosures arrive as discrete and irregular events, but their consequences unfold through continuous financial time series. 
This interaction between discrete web events and continuous market dynamics raises an important problem: how can we estimate the short-term abnormal market response following an event from the market context observed before its disclosure?

This problem is practically important because cybersecurity incidents can trigger abrupt stock price movements and economic losses. 
Empirical studies show that major breaches can erase billions of dollars in market capitalization within days, erode consumer trust, and depress trading activity for weeks. 
At the macroeconomic level, cybercrime is estimated to cause annual global losses on the order of trillions of dollars~\cite{martins2025stock, AKYILDIRIM2024102082}. 
For investors, firms, and regulators, the key quantity of interest is often not the entire post-event price trajectory, which is affected by many confounding factors after disclosure, but the magnitude of the abnormal loss associated with the event. 
Accurately estimating this impact is essential for cyber-risk assessment, portfolio protection, and understanding how online security incidents propagate through financial markets.

\begin{figure}[t]
    \centering
    \vspace{0.1in}
    \includegraphics[width=0.47\textwidth]{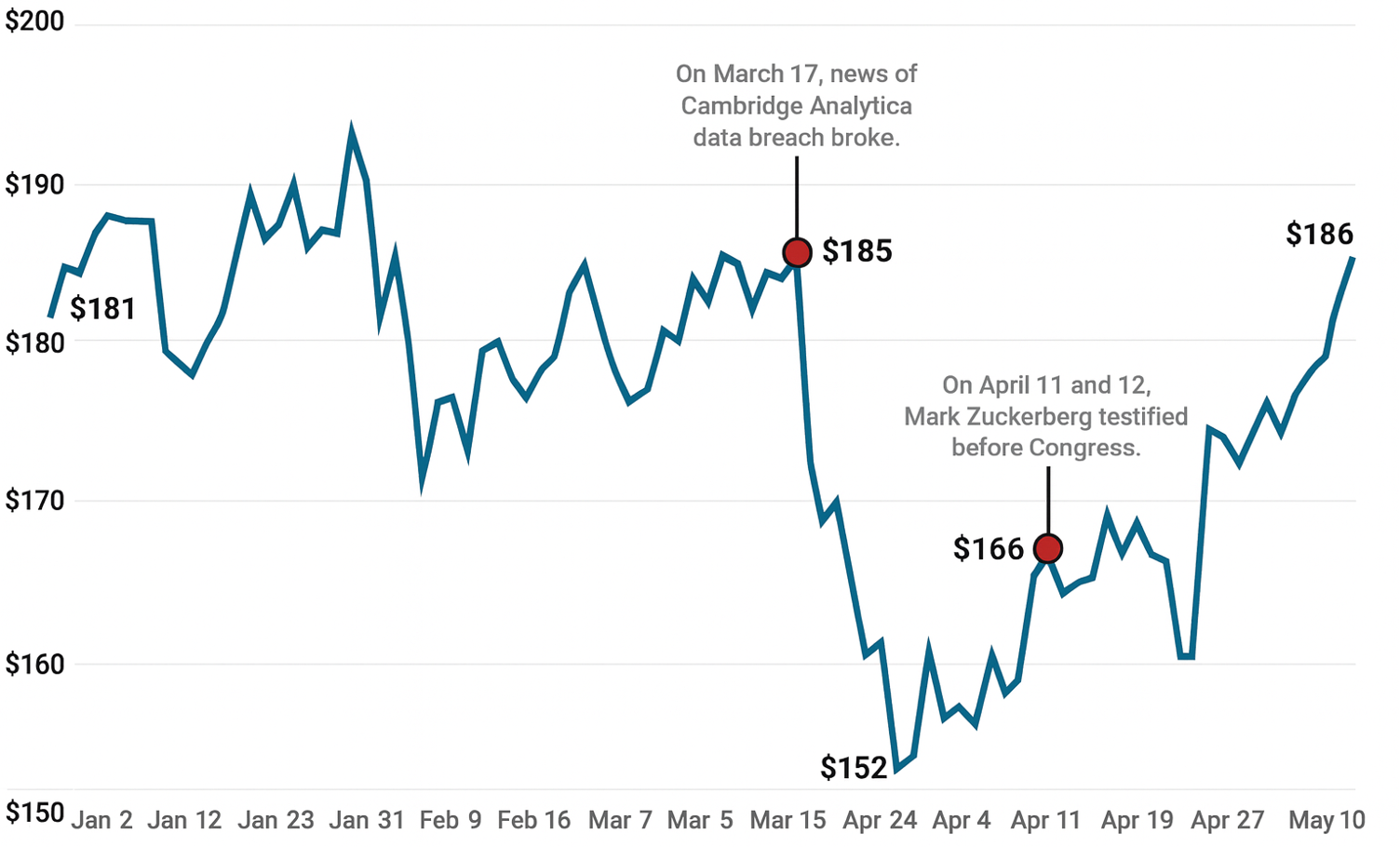} 
    \vspace{-0.1in}
    \caption{Stock market shock induced by a cybersecurity event: Facebook's 2018 data breach.}
    \label{fig:facebook}
\end{figure}

Figure~\ref{fig:facebook} shows this perspective using the Facebook--Cambridge Analytica disclosure in March 2018. 
Following the disclosure, Facebook's stock price dropped sharply from approximately \$185 to \$152 within days, reflecting an immediate market reaction to the cybersecurity-related event~\cite{jeleskovic2024impact}. 
Such examples highlight why event-aware modeling is needed: forecasting systems that rely only on endogenous temporal patterns may underestimate abnormal losses during shocks, while models that ignore the event context cannot distinguish event-driven reactions from routine market volatility. 
Failing to quantify these impacts can lead to mispriced risk, inadequate portfolio protection, and limited regulatory visibility into how shocks spread across firms and industries~\cite{sahiner2024volatility}. 
Therefore, modeling the impact of rare but consequential cybersecurity disclosures is a crucial step toward more resilient and adaptive financial decision-making.



Despite rapid advances in time-series forecasting, most existing models are primarily designed to capture endogenous temporal regularities, such as long-term trends, seasonal cycles, and short-term autocorrelation~\cite{kong2025deep, wen2022transformers, kim2025comprehensive}. 
Discrete external events, such as cybersecurity disclosures, are often underrepresented or incorporated only as auxiliary covariates through simple concatenation or indicator features. 
Such treatments are insufficient for event-conditioned impact prediction, where the goal is not to extrapolate a smooth future trajectory, but to estimate how a rare event interacts with the pre-event market state and induces abnormal loss. 
In particular, existing models often fail to capture which aspects of historical dynamics make a firm more vulnerable to a disclosure, how event metadata should be aligned with temporal representations, and how to learn robust event-aware patterns from sparse observations. 
We therefore identify three key challenges in modeling the impact of discrete events on continuous financial signals:

\begin{itemize}[leftmargin=1.5em]
    \item \textbf{Multi-resolution market context}: 
    Pre-event financial dynamics span multiple temporal scales, including long-term valuation trends, medium-term market regimes, and short-term volatility. 
    These scales jointly shape market responses to cybersecurity disclosures, requiring models to capture multi-resolution context for abnormal impact estimation.

    \item \textbf{Event--temporal integration}: 
    Cybersecurity incidents involve heterogeneous, non-temporal metadata, such as breach type, affected scale, severity, and data sensitivity. 
    Their financial consequences depend on interactions with the market state, which simple feature concatenation cannot explicitly capture.

    \item \textbf{Sparse event supervision}: 
    High-impact cybersecurity disclosures are rare and diverse, yielding limited labeled data. 
    Standard regression may overfit background market patterns and underuse event semantics, motivating training strategies that amplify event-relevant signals and separate similar from dissimilar event--market contexts.
\end{itemize}

To address these challenges, we propose \textbf{EventTime}, a unified framework for event-conditioned financial impact prediction.
EventTime models pre-event dynamics at multiple temporal resolutions to capture long-horizon market context and short-horizon fluctuations.
An attention-based fusion mechanism aligns structured event metadata with temporal signals to identify recent patterns most relevant to the event.
To address sparse supervision from rare and heterogeneous events, EventTime further uses dynamic contrastive learning to strengthen event-relevant representations by contrasting similar and dissimilar event--market contexts.
Together, these components improve short-term abnormal loss estimation over conventional forecasting models that mainly rely on endogenous temporal patterns.

We evaluate EventTime on a real-world dataset that links publicly disclosed cybersecurity incidents with U.S. stock market data. 
This setting provides a challenging testbed for event-conditioned impact prediction: cybersecurity disclosures are sparse and heterogeneous, yet they can induce disproportionate abnormal losses within a short reaction window. 
Experimental results show that EventTime consistently outperforms strong baselines, including state-of-the-art time-series forecasting models and event-augmented variants, in estimating post-event financial losses. 
To examine the broader applicability of our framework, we further evaluate EventTime on a hydrological event-impact prediction task, where extreme rainfall events are used to predict subsequent streamflow responses. 
This cross-domain study shows that the same event-conditioned modeling principle can also capture natural shocks in environmental systems, underscoring the generality of our approach.

In summary, this paper makes the following contributions:
\begin{itemize}[leftmargin=2em]
    \item We formulate cybersecurity event impact prediction as an event-conditioned time-series task, where the objective is to estimate short-term abnormal financial loss from pre-event market dynamics and event metadata, rather than reconstruct the full post-event trajectory.
    
    \item We propose \textbf{EventTime}, combining multi-resolution encoding of pre-event market signals, attention-based fusion of heterogeneous event metadata, and dynamic contrastive learning to strengthen rare-event signals under sparse supervision.
    
    \item We develop a novel dataset, \textbf{SECURE}, by integrating stock-level financial time series from Yahoo Finance\footnote{\url{https://github.com/ranaroussi/yfinance}} with cybersecurity event records from the Privacy Rights Clearinghouse (PRC)~\cite{prc}, enabling a systematic study of how disclosed security incidents affect short-term market losses.
    
    \item We empirically demonstrate that EventTime consistently outperforms competitive time-series and event-aware baselines on the SECURE dataset. Beyond the financial domain, we further validate our approach on a hydrological event-impact prediction benchmark, illustrating its potential to model diverse event-driven responses.
\end{itemize}
\begin{figure*}[t]
\vspace{-0.05in}
    \centering
    \includegraphics[width=0.95\textwidth]{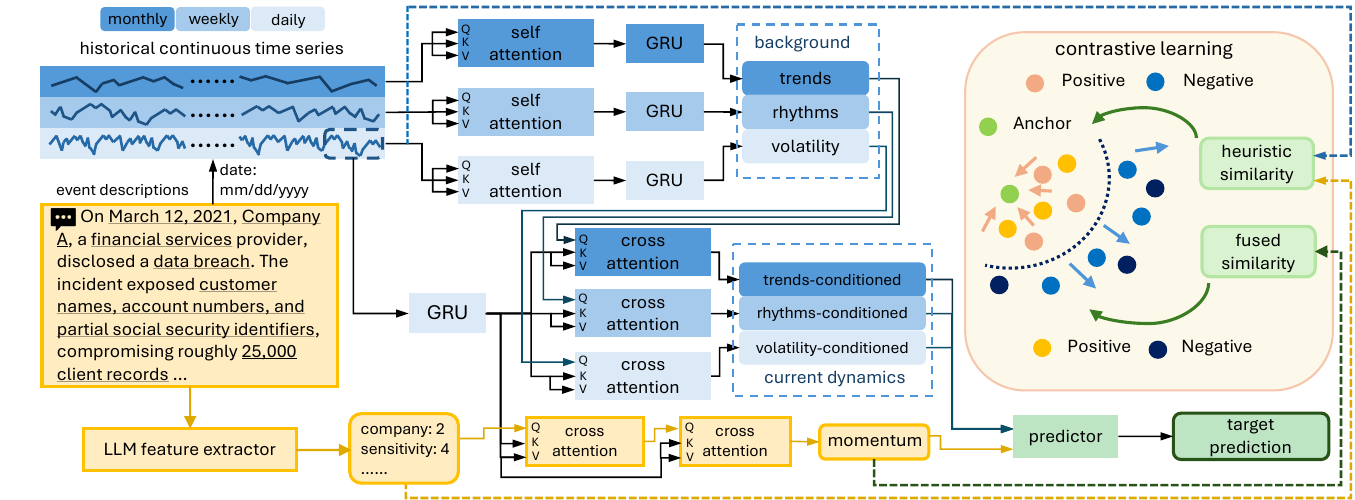} 
    \caption{Overall framework of EventTime.}
    \label{fig:framework}
\vspace{-0.1in}
\end{figure*}
\section{Problem Definition}

We formulate the problem as \emph{event-conditioned impact prediction}. 
Given a historical time-series segment before a discrete external event and metadata describing the event, the goal is to estimate a scalar post-event impact score. 
This formulation differs from conventional multi-step forecasting, which predicts the full future trajectory. 
In our setting, the post-event trajectory contains both event-induced reactions and unrelated market movements, such as market-wide fluctuations, sector trends, and subsequent news. 
Directly forecasting every future time step can therefore obscure the specific effect of the event. 
Instead, we focus on estimating a compact impact label that summarizes the short-term abnormal response associated with the event.

Let \(X = \{x_1, x_2, \dots, x_T\}\) denote a pre-event economic time series, where each observation \(x_t \in \mathbb{R}\) represents a financial indicator, such as excess return, at time \(t\). 
Each sequence segment is constructed around a discrete external event, with the final observed time step \(T\) aligned to the event occurrence or disclosure date. 
The event is characterized by a metadata vector \(e_T \in \mathbb{R}^k\), which includes attributes such as event type, affected scale, incident severity, and data sensitivity.

Rather than assuming that the pre-event sequence has a single characteristic frequency, we consider that \(X\) contains dynamics across multiple temporal scales. 
We therefore define a set of scale-specific transformations \(\{f^{(1)}, f^{(2)}, \dots, f^{(M)}\}\), where each \(f^{(m)}\) extracts patterns of \(X\) at a different granularity, such as long-term trends, medium-term regimes, or short-term fluctuations. 
This yields a collection of multi-scale representations:
\[
\mathcal{X} = \{X^{(m)} = f^{(m)}(X) \mid m = 1, \dots, M\},
\]
where each \(X^{(m)}\) represents a different temporal resolution of the same pre-event sequence.

The prediction target is a scalar outcome $y \in \mathbb{R}$ that summarizes the short-term abnormal response after event disclosure.
The precise definition of \(y\) may vary across domains. 
In our financial setting, \(y\) is instantiated as the maximum excess-return decline within a short post-event window, capturing the worst abnormal loss after disclosure while filtering out market- and sector-wide movements. 
Formally, the objective is to learn a function
\[
\hat{y} = f_\theta(\mathcal{X}, e_T),
\]
where \(f_\theta\) maps pre-event multi-scale time-series dynamics and event metadata to the predicted impact score.

\section{Methodology}

We present \textbf{EventTime}, a unified framework for event-conditioned financial impact prediction. 
Given a segment of market history preceding a disclosed cybersecurity event, together with structured and unstructured metadata describing the event, EventTime estimates the event's short-term abnormal financial impact. 
The framework is built around three core components: 
(i) a \emph{multi-resolution time-series encoder} that captures pre-event market context across multiple temporal scales; 
(ii) an \emph{event--feature fusion module} that aligns heterogeneous event metadata with temporal representations; and 
(iii) a \emph{dynamic contrastive learning strategy} that strengthens rare-event signals under sparse supervision. 
An overview of the architecture is provided in Figure~\ref{fig:framework}.

\subsection{Model Structure}

\subsubsection{Multi-resolution Time-series Encoder}

Pre-event financial dynamics contain information at multiple temporal scales. 
Long-horizon patterns reflect the broader market condition and firm-level trajectory before disclosure, while shorter horizons capture recent volatility, momentum, and local instability. 
These signals jointly shape how the market reacts to a cybersecurity event: the same disclosure may induce different losses depending on whether the firm is already in a stable, volatile, declining, or recovering state. 
Modeling these multi-scale dynamics therefore provides the contextual background for estimating event-attributable impact.

To capture this structure, EventTime employs a multi-resolution long-term encoder that jointly processes daily, weekly, and monthly views of the historical sequence. 
Given a raw daily sequence \(X^{(d)} = \{x_1, \dots, x_T\}\), where \(T\) denotes the event date, we derive aggregated signals by temporal averaging:
\[
X^{(w)}_t = \frac{1}{n_w}\sum_{i=0}^{n_w-1} x_{t-i}, \quad
X^{(m)}_t = \frac{1}{n_m}\sum_{i=0}^{n_m-1} x_{t-i},
\]
where \(n_w\) and \(n_m\) denote the number of trading days in a week and a month, respectively.
This produces aligned views \(\{X^{(d)}, X^{(w)}, X^{(m)}\}\), each emphasizing a different temporal granularity of the same pre-event history.

Each view is encoded by a dedicated branch \(E^{(z)}(\cdot)\), \(z \in \{d,w,m\}\), producing
\[
H^{(z)} = E^{(z)}(X^{(z)}), \quad H^{(z)} \in \mathbb{R}^{p}.
\]
The daily encoder \(E^{(d)}(\cdot)\) models fine-grained fluctuations, the weekly encoder \(E^{(w)}(\cdot)\) captures medium-term patterns, and the monthly encoder \(E^{(m)}(\cdot)\) summarizes coarse-grained market context. 
Together, the embeddings \(\{H^{(d)}, H^{(w)}, H^{(m)}\}\) form a multi-resolution representation of the pre-event market state, which is later used for event-conditioned impact estimation.

In addition to the long-term multi-resolution context, EventTime includes a short-term encoder that focuses on the most recent daily window before the event. 
While the long-term branches summarize broader historical conditions, the short-term encoder isolates the local market state immediately preceding disclosure. 
Concretely, from the daily sequence \(X^{(d)} = \{x_1, \dots, x_T\}\), we extract a local window of length \(L\):
\[
X^{(s)} = \{x_{T-L+1}, \dots, x_T\}.
\]
This window is processed by a short-term encoder \(E^{(s)}(\cdot)\), producing time-step-level embeddings
\[
H^{(s)} = E^{(s)}(X^{(s)}), \quad H^{(s)} \in \mathbb{R}^{L \times p}.
\]
This design reflects the intuition that event impact depends on both broad historical context and immediate pre-event conditions: the former characterizes baseline vulnerability or resilience, while the latter captures the local state with which the event interacts.

To integrate long-term context with local momentum, EventTime performs cross-attention between the two representations. 
For each long-term embedding \(H^{(z)} \in \mathbb{R}^{1 \times p}\), \(z \in \{d,w,m\}\), we use \(H^{(z)}\) as the query, while the short-term embedding \(H^{(s)}\) provides the keys and values:
\[
\tilde{H}^{(z)} = \mathrm{Attn}\!\left(
Q=H^{(z)}W_Q^{(z)},\;
K=H^{(s)}W_K^{(z)},\;
V=H^{(s)}W_V^{(z)}
\right),
\]
where \(W_Q^{(z)}, W_K^{(z)}, W_V^{(z)}\) are learnable projection matrices. 
This operation allows each scale-specific long-term summary to selectively attend to recent observations that are most relevant to that scale. 
The resulting representations \(\{\tilde{H}^{(d)}, \tilde{H}^{(w)}, \tilde{H}^{(m)}\}\) preserve multi-resolution historical context while emphasizing salient pre-event fluctuations for subsequent event-conditioned fusion.

\subsubsection{Event Fusion Module}

While multi-resolution and short-term encoders capture endogenous dynamics, financial time series are also influenced by exogenous shocks. Firm-specific disclosures, such as cybersecurity incidents or earnings announcements, can induce abnormal returns and volatility spikes that deviate from historical patterns. Ignoring such events degrades predictive accuracy by conflating event-driven losses with routine fluctuations. Therefore, explicitly incorporating structured event metadata alongside temporal signals is essential for modeling event-conditioned momentum.

We define this momentum not only as recent price fluctuations, but as fluctuations conditioned on the event at time $T$. Formally, let $e_T \in \mathbb{R}^{p_m}$ denote the event metadata (e.g., type, severity, affected customers), which is projected by an event encoder
$H^{(e)} = E^{(e)}(e_T) \in \mathbb{R}^p.$ 
A naive solution would be to directly concatenate $H^{(e)}$ with the short-term embeddings $H^{(s)} \in \mathbb{R}^{L \times p}$. However, such early fusion overlooks the heterogeneity of daily signals: some pre-event days already reflect anticipatory volatility driven by rumors, sector contagion, or correlated shocks, while others capture only routine fluctuations. The coupling between daily dynamics and the event therefore varies as markets internalize indirect signals even before disclosure. Treating them equally risks diluting the true event-driven signal.

We introduce an event fusion module where event queries the short-term sequence to identify sensitive positions and modulate their influence through a two-layer attention mechanism, interpretable as a translate–locate–imprint process.

\paragraph{(1) Translate (Event-to-Temporal Projection).}
A key challenge is the heterogeneous nature of event metadata and temporal observations. 
In the first attention layer, the event embedding $H^{(e)}$ serves as the query while the short-term daily embeddings $H^{(s)}$ act as keys and values, projecting the event into temporal modality:
{
\[
\alpha = \frac{(H^{(e)}W_q^{(t)})(H^{(s)}W_k^{(t)})^\top}{\sqrt{d}} \in \mathbb{R}^{1 \times L},~c = \alpha \, H^{(s)} W_v^{(t)} \in \mathbb{R}^{1 \times p}.
\]}
Here $\alpha$ highlights event-relevant temporal positions and $c$ is a translated event representation in the temporal modality.

\paragraph{(2) Locate (Event-Guided Re-attention).}
In the second attention layer, the translated summary $c \in \mathbb{R}^{1 \times p}$ serves as the query to re-attend over the short-term sequence $H^{(s)}$ (as keys and values), producing contextual weights
\[
\beta = \frac{(c W^{(l)}_q)(H^{(s)} W^{(l)}_k)^\top}{\sqrt{d}}\in \mathbb{R}^{1 \times L}.
\]
The distribution $\beta$ highlights which pre-event positions are most related under the event context.

\paragraph{(3) Imprint (Contextual Modulation).}
Finally, the attention weights $\beta$ are imprinted back onto the short-term tokens to yield an event-aware sequence representation:
{\footnotesize

\[
\tilde{H}^{(s)} = H^{(s)} + \beta^\top H^{(e)} \in \mathbb{R}^{L \times p}.
\]
}

This modulation reinforces abnormal fluctuations that align with the event while down-weighting routine noise, producing $\tilde{H}^{(s)}$ as an event-conditioned refinement of short-term momentum. Finally, the momentum $\tilde H^{(s)}$ is passed through the fusion encoder $E^{(f)}$ to obtain the final fused representation:
$H^{(f)} = E^{(f)}(\tilde H^{(s)}) \in \mathbb{R}^{p}$.

In summary, the event fusion module maps event metadata to time series, finds the most relevant pre-event points, and blends the event effect into local dynamics. 
The fused representation \(H^{(f)}\) keeps the event information and short-term context, and is then combined with the long-term embeddings \(\{\tilde{H}^{(d)}, \tilde{H}^{(w)}, \tilde{H}^{(m)}\}\) to produce a single combined representation to predict the impact of the post-event.


To support this multi-resolution modeling, our architecture includes three specialized components: a Transformer encoder~\cite{vaswani2017attention} followed by a GRU module~\cite{dey2017gate} to capture long-term dependencies across daily, weekly, and monthly sequences; a GRU-based encoder to summarize short-term patterns before the event; and a multi-layer perceptron (MLP)~\cite{riedmiller2014multi} that fuses the event-conditioned short-term representation with long-term context. Figure~\ref{fig:framework} provides an overview of the overall architecture and information flow.

\subsection{Dynamic Contrastive Learning}

While supervised objectives are necessary for event-conditioned impact prediction, they are often insufficient in financial settings. 
Labeled instances are scarce because only a limited number of cyberattack disclosures can be reliably aligned with publicly traded firms and valid market histories. 
Moreover, impact labels are noisy: post-event losses may reflect not only the disclosed incident, but also market-wide fluctuations, sector trends, and unrelated firm-specific news. 
A regression loss encourages accurate fitting of scalar impact values, but provides limited guidance on how the representation space should be organized with respect to event semantics and event--market context. 
As a result, the model may rely heavily on endogenous time-series signals, while the sparse but informative event metadata receives weak supervision.

To address this limitation, we introduce a dynamic contrastive learning objective as auxiliary representation-level supervision. 
The key intuition is that two samples should have closer representations if they share similar event--market contexts, even when their exact impact labels are noisy. 
For example, two disclosures with similar breach severity, affected scale, data sensitivity, and pre-event volatility patterns may induce comparable market responses, whereas an unrelated disclosure under a different market condition should be separated in the latent space. 
This pairwise supervision complements the pointwise regression loss by encouraging the model to preserve event-relevant structure beyond what can be learned from individual labels.

A central challenge is how to define positive and negative pairs. 
Using only event metadata may ignore the fact that the same type of cybersecurity incident can lead to different financial impacts depending on the firm's pre-event market state. 
Conversely, using only time-series similarity may group together firms with similar price movements but unrelated event semantics. 
We therefore compute sample similarity from two complementary views: a heuristic view based on stable pre-computed event and time-series descriptors, and a learned view based on the fused event-conditioned representations produced by EventTime. 
The heuristic view stabilizes pair construction early in training, when model embeddings are not yet reliable, while the learned view gradually captures task-specific event--temporal relationships as training progresses. 
This dynamic design allows contrastive learning to evolve from prior-guided alignment to model-driven alignment.

\subsubsection{Similarity Computation}

To enable contrastive learning, we measure similarity using both heuristic statistics and learned representations. 
For numerical stability, we define a normalized cosine similarity
\[
\mathrm{sim}(a,b)=\frac{1+\cos(a,b)}{2},
\]
which maps cosine similarity to the range \([0,1]\).

\paragraph{Heuristic similarity.}
From each short-term pre-event window, we compute a descriptive feature vector \(T^{(s)}\) summarizing key aspects of market behavior:
\textit{central tendency and dispersion} (mean and standard deviation),
\textit{extreme values} (maximum and minimum),
\textit{temporal dynamics} (linear slope of returns),
\textit{volatility change} (variance difference between early and late segments),
and \textit{recent shocks} (relative change on the last day and over the last five days).
These statistics provide a compact summary of the local market state immediately before the event, capturing both stable patterns and abrupt fluctuations.

For the event side, we use the metadata embedding \(e_T\) described in Section~\ref{dataset}, which encodes attributes such as breach type, incident severity, affected scale, and data sensitivity. 
Similarity between event metadata vectors reflects event-level resemblance. 
The heuristic similarity between samples \(i\) and \(j\) is defined as
\[
S^{\text{heur}}_{ij} =
\big[\mathrm{sim}(T^{(s)}_i, T^{(s)}_j)\big]^{\alpha}
\cdot
\big[\mathrm{sim}(e_{T_i}, e_{T_j})\big]^{1-\alpha},
\]
where \(\alpha \in [0,1]\) balances the influence of short-term time-series statistics and event metadata. 
This score favors pairs that are similar in both pre-event market condition and event semantics.

\paragraph{Learned similarity.}
Heuristic descriptors are stable but limited, as they may miss nonlinear interactions between event attributes and temporal dynamics. 
We therefore also compute similarity from the fused event-conditioned embeddings \(H^{(f)}\) obtained by the event fusion module:
\[
S^{\text{fuse}}_{ij} = \mathrm{sim}\!\left(H^{(f)}_i, H^{(f)}_j\right).
\]
This learned similarity evolves during training and captures task-specific relationships that heuristic features may overlook.

\paragraph{Dynamic weighting.}
In the early stages of training, when encoder representations are not yet well formed, we place greater emphasis on heuristic similarity. 
As the model improves, we gradually increase the contribution of learned similarity. 
Formally, the combined similarity at epoch \(t\) is
\[
S_{ij}(t) =
\big(S^{\text{heur}}_{ij}\big)^{1-\lambda_t}
\cdot
\big(S^{\text{fuse}}_{ij}\big)^{\lambda_t},
\]
where \(\lambda_t \in [0,1]\) is a scheduling coefficient that increases with training progress. 
This dynamic mixture ensures stable pair construction at the beginning of training while progressively shifting toward more expressive, model-driven similarities.

\paragraph{Contrastive pair selection.}
For each anchor sample \(i\), positive samples are selected from the top-\(k\) most similar samples whose dynamic similarity \(S_{ij}(t)\) exceeds a threshold \(\rho\). 
Negative samples are selected from samples whose similarity is below a lower threshold \(\rho'\). 
This thresholded top-\(k\) strategy avoids treating all non-positive samples as equally negative, which is important because different cybersecurity events may share partial semantic or temporal similarity. 
The resulting pairs provide auxiliary supervision that encourages the representation space to reflect event--market context rather than only scalar label proximity.

\subsubsection{Training Objectives}

Our model is trained using a combination of supervised prediction and auxiliary contrastive objectives. 
The prediction objective estimates the post-event impact \(y\) from the fused long- and short-term representations \(\{\tilde{H}^{(d)}, \tilde{H}^{(w)}, \tilde{H}^{(m)}\}\) together with the event-fused summary \(H^{(f)}\), optimized using the mean squared error loss
\[
\mathcal{L}_{\text{pred}} = \|y-\hat{y}\|_2^2.
\]

For contrastive learning, the dynamic similarity \(S_{ij}(t)\) is used to construct positive and negative pairs, while the contrastive logits are computed from the learned fused representations. 
Specifically, for each anchor \(i\), let \(Q_i^+\) and \(Q_i^-\) denote its positive and negative sets. 
We define the representation similarity
\[
\hat{S}^{\text{fuse}}_{ik}=\mathrm{sim}\!\left(H^{(f)}_i,H^{(f)}_k\right),
\]
and optimize the InfoNCE-style loss
\[
\mathcal{L}_{\text{contrast}}
=
-\sum_i
\log
\frac{
\sum_{j\in Q_i^+}\exp\!\left(\hat{S}^{\text{fuse}}_{ij}/\tau\right)
}{
\sum_{k\in Q_i^+\cup Q_i^-}\exp\!\left(\hat{S}^{\text{fuse}}_{ik}/\tau\right)
},
\]
where \(\tau\) is a temperature parameter. 
Using \(S_{ij}(t)\) for pair construction and \(\hat{S}^{\text{fuse}}_{ij}\) for contrastive logits separates the mining criterion from the optimized representation similarity, making the objective more stable.

To further align learned representations with stable heuristic priors, we introduce a similarity regularization term:
\[
\mathcal{L}_{\text{similarity}}
=
\sum_{i,j}
\left(
\hat{S}^{\text{fuse}}_{ij}
-
S^{\text{heur}}_{ij}
\right)^2.
\]
This term prevents the learned similarity from drifting too far from event and market descriptors that are known before training, while the dynamic contrastive objective still allows the model to discover task-specific relationships.

The final training objective is
\[
\mathcal{L}
=
\mathcal{L}_{\text{pred}}
+
\lambda \mathcal{L}_{\text{contrast}}
+
\gamma \mathcal{L}_{\text{similarity}},
\]
where \(\lambda\) and \(\gamma\) balance the auxiliary losses. 
This joint objective encourages accurate impact prediction while structuring the embedding space so that rare but high-impact events exert stronger influence on representation learning.

\section{Experiments}

    
    
    
To evaluate the effectiveness and interpretability of our approach, we conduct extensive experiments on the SECURE dataset centered around four research questions.
\begin{itemize}[leftmargin=2em]
\item \textbf{RQ1 (Model Effectiveness)} evaluates how EventTime compares with strong time-series and event-aware baselines in estimating post-event abnormal financial loss.

\item \textbf{RQ2 (Training Strategy)} examines whether dynamic contrastive learning improves performance beyond standard supervised fine-tuning and how it enhances the quality of learned event-conditioned representations.

\item \textbf{RQ3 (Event Fusion Design)} investigates whether the proposed two-layer \textit{translate--locate} attention mechanism improves event--temporal fusion over a single-layer design and identifies which event features most strongly influence the learned impact representation.

\item \textbf{RQ4 (Model Robustness)} assesses whether contrastive learning increases resilience to missing event metadata.
\end{itemize}

\subsection{Dataset}
\label{dataset}

We construct SECURE (\underline{S}tock-market \underline{E}vent-driven dataset for \underline{C}yber breach \underline{U}nderstanding and \underline{R}esponse \underline{E}stimation) by aligning stock-level financial time series with firm-level cybersecurity disclosures. 
Daily closing prices and sector indices are obtained from Yahoo Finance, while cybersecurity incidents are collected from the Privacy Rights Clearinghouse (PRC), which documents data breaches, ransomware attacks, and related security events with attributes such as affected firm, disclosure date, and number of customers impacted. 
SECURE covers 268 publicly listed firms from 2005 to 2025 and 3,034 documented incidents. 
After filtering for valid public-firm matches and sufficient financial history, the final dataset contains 1,128 firm--event pairs, with an average of approximately four events per firm.

For each affected firm, we align the input sequence so that the final observed day corresponds to the disclosure date \(T\). 
This ensures that the model only observes pre-event market dynamics when estimating post-event impact. 
We use a long-term window of \(L_{\mathrm{long}}=520\) trading days, approximately two years, and a short-term window of \(L_{\mathrm{short}}=20\) trading days, approximately one month. 
The input sequence \(X=\{x_1,\dots,x_T\}\) consists of daily excess returns,
\[
x_t=
\frac{P_t-P_{t-1}}{P_{t-1}}
-
\frac{IX_t-IX_{t-1}}{IX_{t-1}},
\]
where \(P_t\) denotes the stock closing price and \(IX_t\) denotes the corresponding industry index. 
This adjustment removes market- and sector-wide movements, thereby emphasizing firm-specific dynamics that are more directly related to the cybersecurity event.

Following standard event-study practice, we define the post-event impact using the maximum excess-return decline within \(d\) trading days after the disclosure:
\[
Y_0=
\min_{T<t\le T+d}
\left[
\frac{P_t-P_T}{P_T}
-
\frac{IX_t-IX_T}{IX_T}
\right],
\]
where \(d=5\), corresponding to one trading week. 
This label summarizes the worst short-term abnormal return after the event, rather than requiring the model to reconstruct the full post-event price trajectory. 
A smaller value of \(Y_0\) indicates a larger abnormal loss. 
In implementation, we use this scalar impact label as the prediction target, consistent with the objective of estimating event-attributable financial loss.

Each event is associated with structured PRC metadata, including \emph{organization type}, \emph{breach type}, and \emph{customers affected}, which provide categorical and numerical context. 
We further process textual descriptions with an LLM-assisted annotation pipeline using GPT-4.1 and a fixed prompt to extract three semantic attributes from disclosure-time information: \emph{incident severity}, an ordinal estimate of event seriousness inferred from both textual reports and structured metadata; \emph{data sensitivity}, which characterizes the type of compromised information, such as Social Security Numbers versus names or addresses; and \emph{company size}, a coarse classification of firm scale. 
We manually inspect a subset of the generated annotations for consistency, and no post-event market information is used in this process.
Together, these features form an event metadata vector \(e_T\in\mathbb{R}^k\) used by the event fusion module.

We split the dataset chronologically according to disclosure dates: the earliest 80\% of firm--event pairs are used for training, and the most recent 20\% are held out for testing. 
This temporal split better reflects a realistic forecasting scenario, where models are trained on historical cybersecurity disclosures and evaluated on future events. 
The split is not constrained to keep firms shared across training and testing; therefore, some test events may correspond to firms that do not appear in the training set. 
This setting makes the evaluation more challenging and tests whether the model can generalize to future disclosures beyond firms observed during training. 
For each event, all input features are constructed only from the pre-disclosure window, and the post-event prices are used solely to compute the scalar impact label.

\begin{figure*}[t]
    \centering
    \begin{subfigure}[t]{0.32\textwidth}
        \centering
        \includegraphics[width=\textwidth]{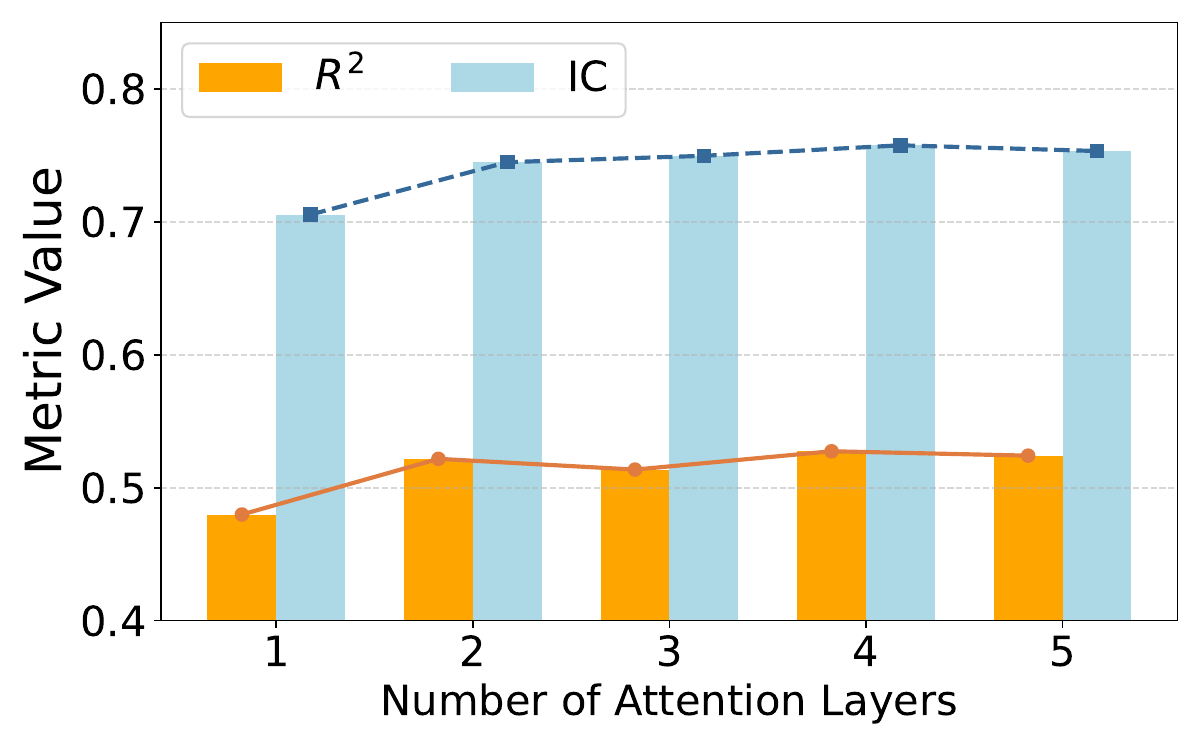}
        \caption{Effect of attention depth in event fusion on downstream prediction performance.}
        \label{fig:attention_depth}
    \end{subfigure}
    \hfill
    \begin{subfigure}[t]{0.32\textwidth}
        \centering
        \includegraphics[width=\textwidth]{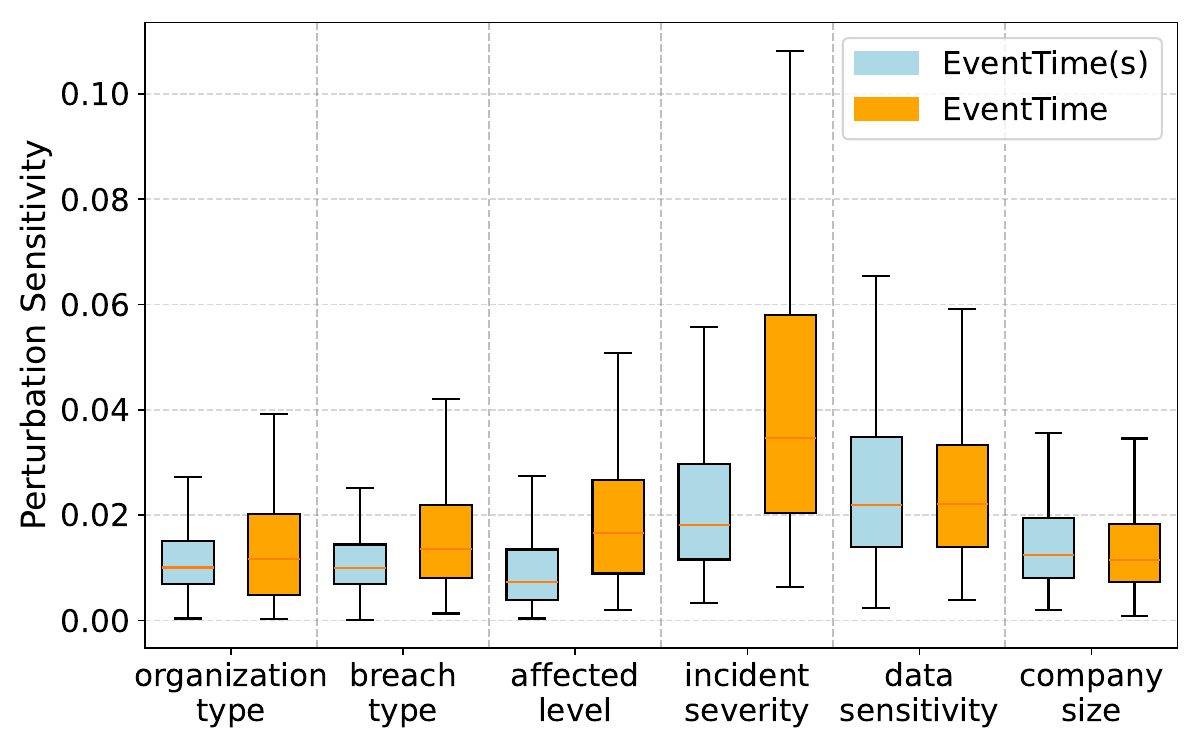}
        \caption{Feature-wise embedding sensitivity of EventTime(s) and EventTime.}
        \label{fig:sensitivity}
    \end{subfigure}
    \hfill
    \begin{subfigure}[t]{0.32\textwidth}
        \centering
        \includegraphics[width=\textwidth]{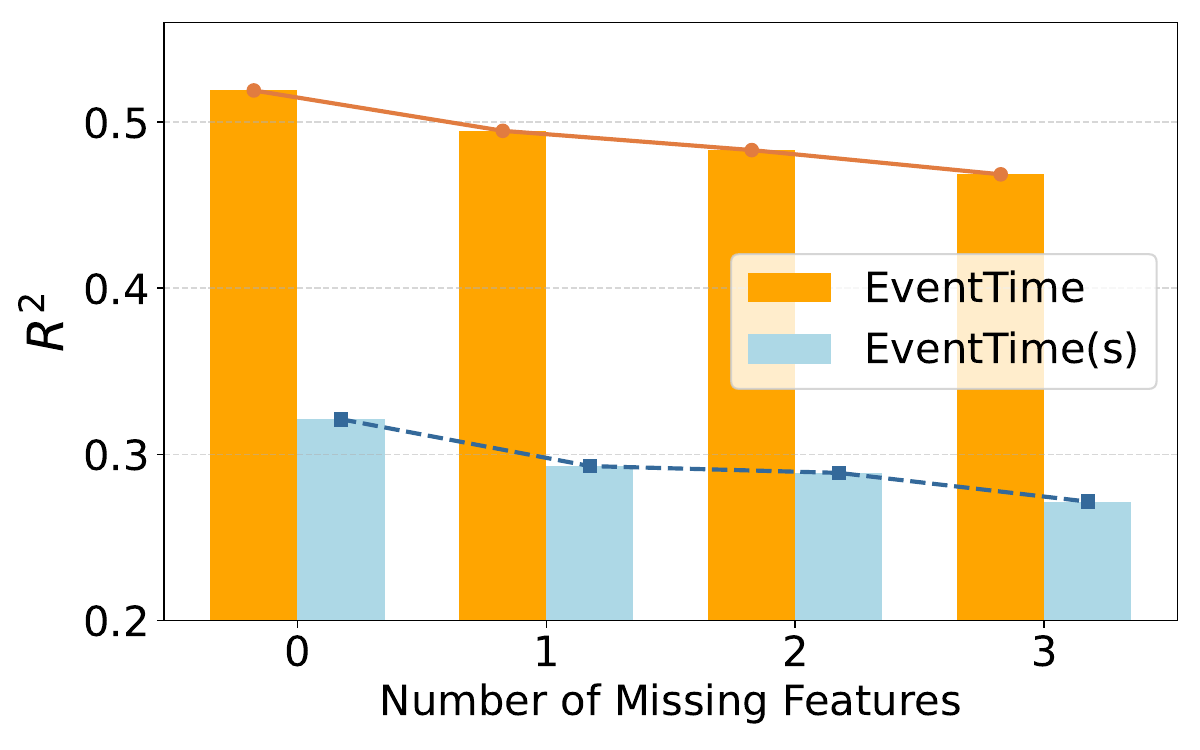}
        \caption{Robustness under increasing event-feature missing rates.}
        \label{fig:missing}
    \end{subfigure}
    \caption{Analysis of EventTime model performance and robustness.}
    \label{fig:case1}
\end{figure*}

\subsection{Experimental Setup}

We compare EventTime with a broad set of representative time-series and event-aware baselines under the same event-conditioned impact prediction setting. 
For all methods, the input consists of the same pre-disclosure financial history and the same event metadata, and the output is the scalar post-event impact label defined in Section~\ref{dataset}. 
Since most time-series forecasting models are not originally designed to incorporate discrete event metadata, we adapt them using a unified late-fusion protocol for fair comparison. 
Specifically, each baseline first encodes the pre-event input window into a time-series representation, while the event metadata is processed using the same event encoder as EventTime. 
The two representations are then concatenated and passed to a shared prediction head to estimate the post-event abnormal loss. 
This design ensures that performance differences mainly reflect the ability of each backbone to model pre-event temporal dynamics, rather than differences in event encoder capacity or prediction-head design.

Our time-series baselines cover several major families of modern forecasting architectures. 
Linear and decomposition-based models include DLinear~\cite{Zeng2022AreTE}, which provides a strong simple baseline for trend-based temporal modeling. 
Mixer-style models include TimeMixer~\cite{wang2023timemixer} and WPMixer~\cite{murad2025wpmixer}, which capture multi-scale temporal patterns through mixing operations. 
Transformer-based baselines include Autoformer~\cite{wu2021autoformer}, PatchTST~\cite{nie2023patchtst}, iTransformer~\cite{liu2024itransformer}, Pyraformer~\cite{liu2022pyraformer}, and TimeXer~\cite{wang2024timexer}, covering decomposition-based attention, patch-based modeling, inverted attention, pyramidal attention, and exogenous-variable-aware forecasting. 
These models provide strong comparisons for assessing whether EventTime's multi-resolution and event-conditioned design offers advantages beyond generic temporal representation learning.

To further assess event awareness, we include two recent event-driven methods. 
Text2Timeseries~\cite{kurisinkel2024text2timeseries} aligns textual event descriptions with time-series dynamics, providing a relevant comparison for models that use event semantics. 
EventTSF~\cite{ge2025eventtsf} applies diffusion-based modeling to generate event-conditioned trajectories. 
Because our task predicts a scalar post-event impact rather than a full future trajectory, we adapt EventTSF by using its event-conditioned representation for scalar impact prediction under the same prediction head. 
All methods are trained with identical input windows, event metadata, train/test splits, and evaluation metrics. 

All methods are trained under the same experimental protocol. 
We use the Adam optimizer with a learning rate of \(10^{-3}\), batch size \(32\), and train each model for \(50\) epochs. 
We set the hidden dimension to \(32\) and use \(4\) attention heads for Transformer-based models when applicable. 
The event encoder is implemented as a two-layer MLP with hidden dimension \(64\), and the final prediction head is kept the same across adapted baselines to ensure fair comparison. 
All reported results are evaluated on the chronological test split described in Section~\ref{dataset}.

We evaluate impact prediction performance using five complementary metrics. 
\(R^2\) measures explanatory power, while MAE and RMSE quantify prediction errors, with RMSE penalizing large mistakes more heavily. 
We also report IC and Rank IC, computed as Pearson and Spearman correlations between predicted and ground-truth impacts, respectively. 
These metrics jointly assess numerical accuracy and ranking consistency, which is important for financial risk screening when exact loss magnitudes are noisy.


\subsection{Quantitative Evaluation}

\subsubsection{Impact Prediction Performance (RQ1)}

To assess the contribution of EventTime’s architecture, we evaluate several backbone baselines alongside EventTime on the SECURE dataset under a uniform fine-tuning setup.
Table~\ref{tab:backbone} summarizes the results across baselines. In our observations, linear baselines such as DLinear perform the worst, confirming that simple trend–seasonality separation is inadequate for event-driven forecasting. Mixer-style models (TimeMixer, WPMixer) offer modest gains but still struggle to capture the joint dependencies between events and temporal dynamics. Transformer architectures perform substantially better, especially those aligned with our motivation: Pyraformer, with its pyramidal multi-scale attention, and TimeXer, which explicitly models exogenous variables, lead among Transformer baselines. Regarding event-centric methods, \textbf{Text2Timeseries} achieves strong performance, demonstrating the effectiveness of aligning textual event semantics with temporal dynamics. In contrast, \textbf{EventTSF} shows unstable behavior, with negative $R^2$ values. We attribute this to its reliance on diffusion-based full-trajectory generation, which is not well suited for the scalar impact prediction required in this setting.

Ultimately, our EventTime(s) variant, which uses the EventTime encoder but trained using only $\mathcal{L}_{\text{pred}}$ (without contrastive and similarity losses), surpasses all of them. These results demonstrate that combining multi-resolution temporal modeling with event-aware fusion, even without contrastive learning, yields a stronger backbone than existing designs and underscores the architectural soundness of EventTime independent of its training strategy.

\begin{table}
\small
    \caption{Comparison of Backbones Under Identical Finetuning Strategy.}
    \label{tab:backbone}
    \centering
    \resizebox{0.47\textwidth}{!}{
    \begin{tabular}{cccccc}
        \toprule
        Model & $R^2$ & MAE & RMSE & IC & Rank IC\\ \midrule
        DLinear & -3.5240 & 0.0315 & 0.0395 & 0.1285 & 0.0716 \\ \midrule
        TimeMixer & 0.0154 & 0.0125 & 0.0160 & 0.1850 & 0.2570 \\ 
        WPMixer & 0.1125 & 0.0121 & 0.0152 & 0.3910 & 0.3855 \\ \midrule
        Autoformer & 0.1450 & 0.0122 & 0.0149 & 0.4310 & 0.2185 \\ 
        PatchTST & 0.2150 & 0.0120 & 0.0143 & 0.3655 & 0.2980 \\ 
        iTransformer & 0.2280 & 0.0118 & 0.0142 & 0.4320 & 0.3750 \\ 
        Pyraformer & 0.2715 & 0.0118 & 0.0138 & 0.5115 & 0.3320 \\ 
        TimeXer & 0.2690 & 0.0119 & 0.0138 & 0.4750 & 0.4765 \\ \midrule
        Text2Timeseries & 0.2854 & 0.0118 & 0.0136 & 0.5162 & 0.4850 \\ \midrule
        EventTime(s) & 0.3212 & 0.0117 & 0.0135 & 0.5145 & 0.4999 \\ \bottomrule
    \end{tabular}}
\end{table}

\begin{table}
\small
    \caption{Comparison of Training Strategies for EventTime. We denote EventTime(s) as the variant trained with supervised loss only, EventTime(-) as the variant using heuristic contrastive similarity, and EventTime as the full model.}
    \label{tab:training}
    \centering
    \resizebox{0.47\textwidth}{!}{
    \begin{tabular}{cccccc}
        \toprule
        Training & $R^2$ & MAE & RMSE & IC & Rank IC\\ \midrule
        EventTime w/o $e$ & 0.2455 & 0.0115 & 0.0145 & 0.4850 & 0.4950 \\ 
        EventTime(s) & 0.3212 & 0.0117 & 0.0135 & 0.5145 & 0.4999 \\ \midrule
        EventTime(-) & 0.4152 & 0.0102 & 0.0128 & 0.6580 & 0.5525\\ 
        EventTime & 0.5242 & 0.0096 & 0.0119 & 0.7446 & 0.5579\\ 
        \bottomrule
    \end{tabular}}
\end{table}

\begin{figure*}[t]
    \centering
    \includegraphics[width=0.32\textwidth]{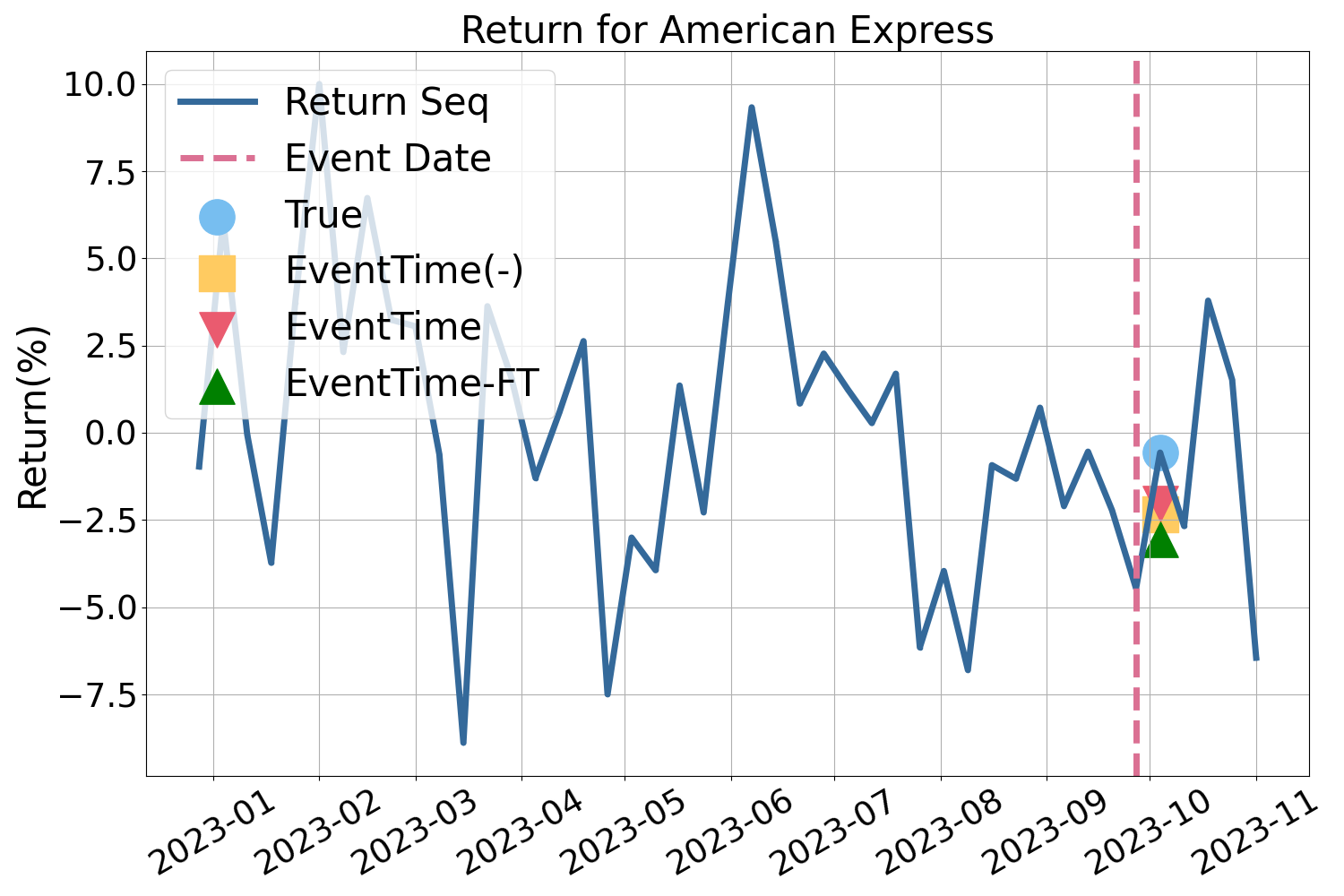}
    \includegraphics[width=0.32\textwidth]{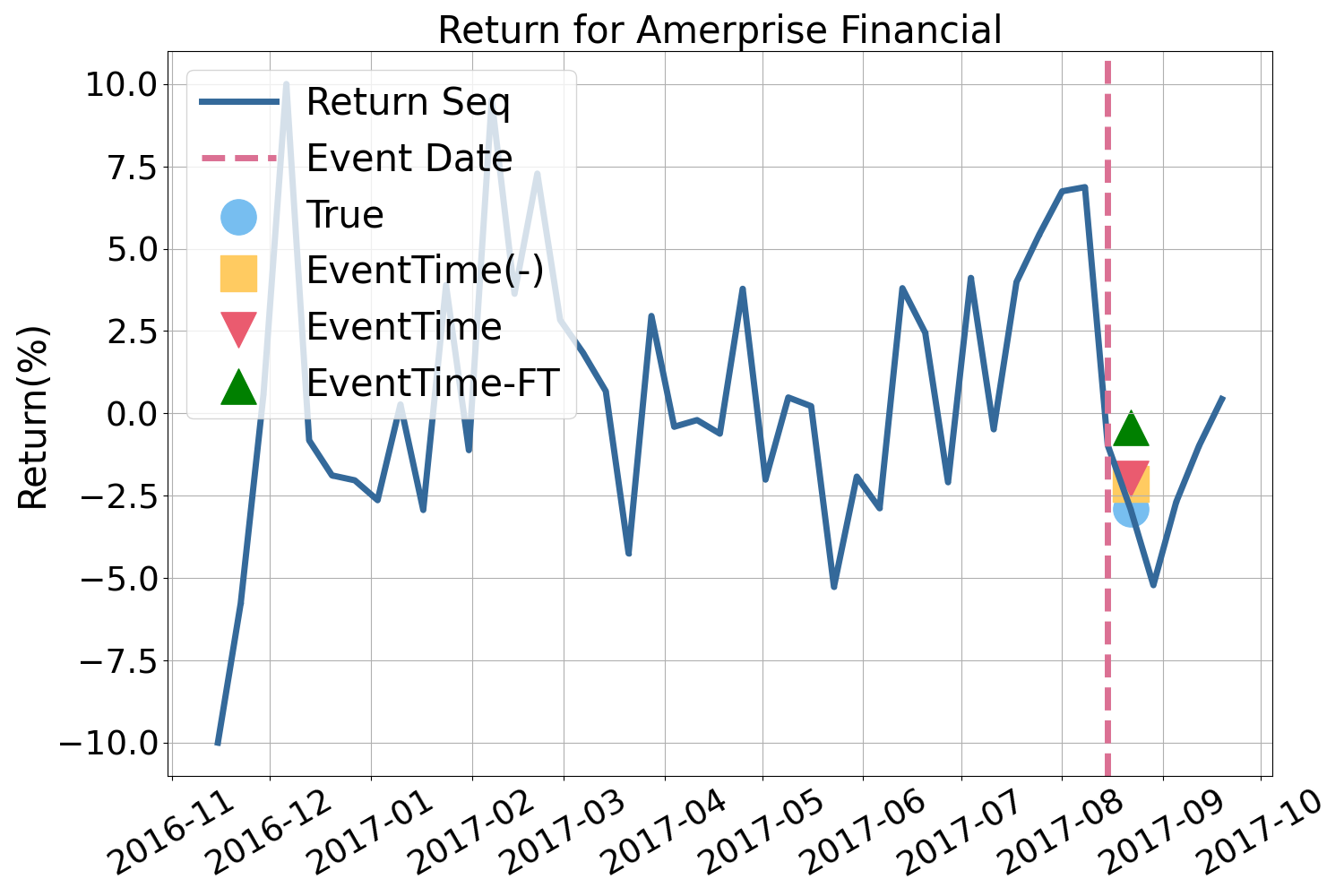}
    \includegraphics[width=0.32\textwidth]{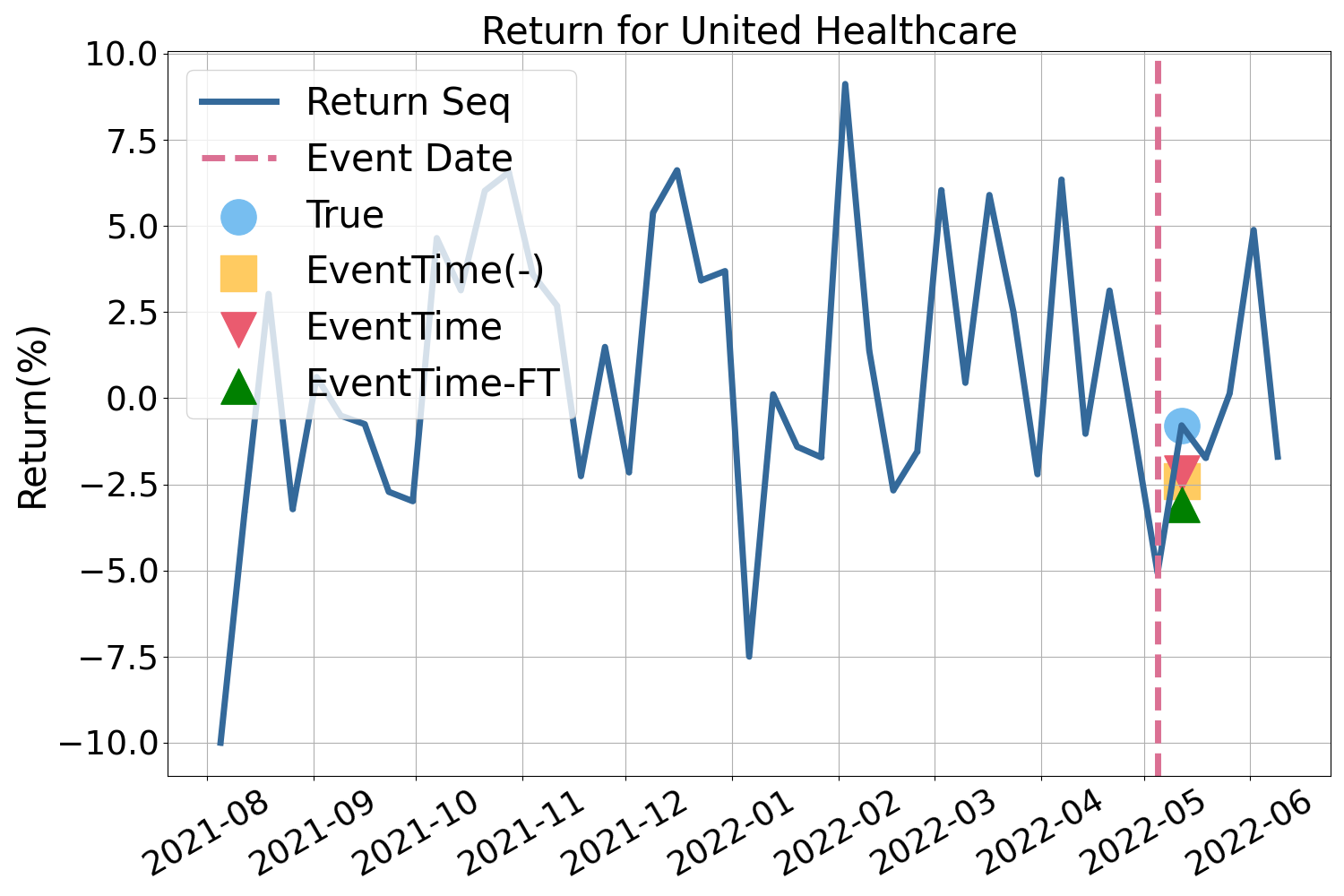}
    \caption{Post-disclosure excess-return dynamics. The target impact label is defined as the maximum excess-return decline.}
    \label{fig:return_series}
\end{figure*}

\begin{figure*}[t]
    \centering
    \includegraphics[width=0.32\textwidth]{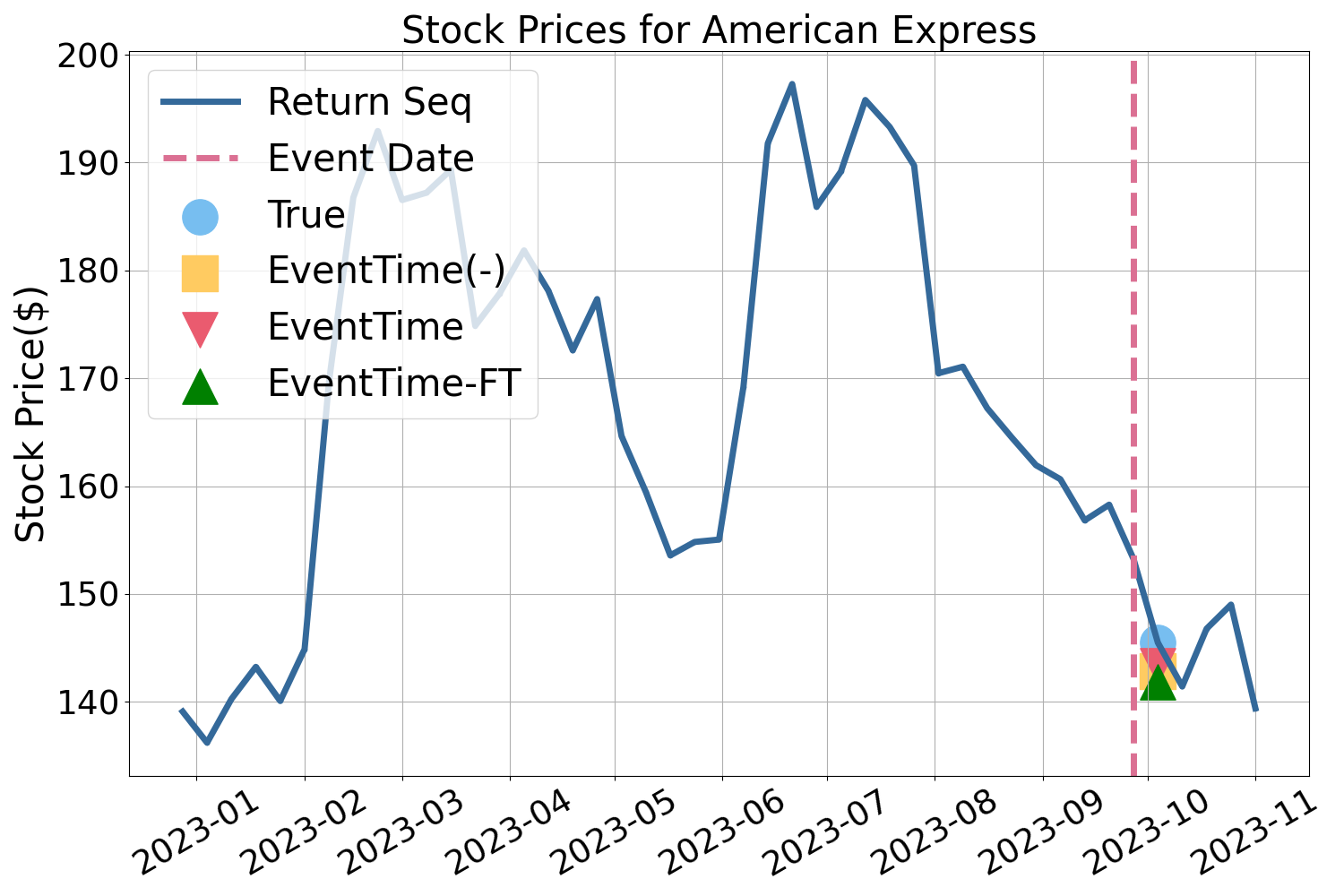}
    \includegraphics[width=0.32\textwidth]{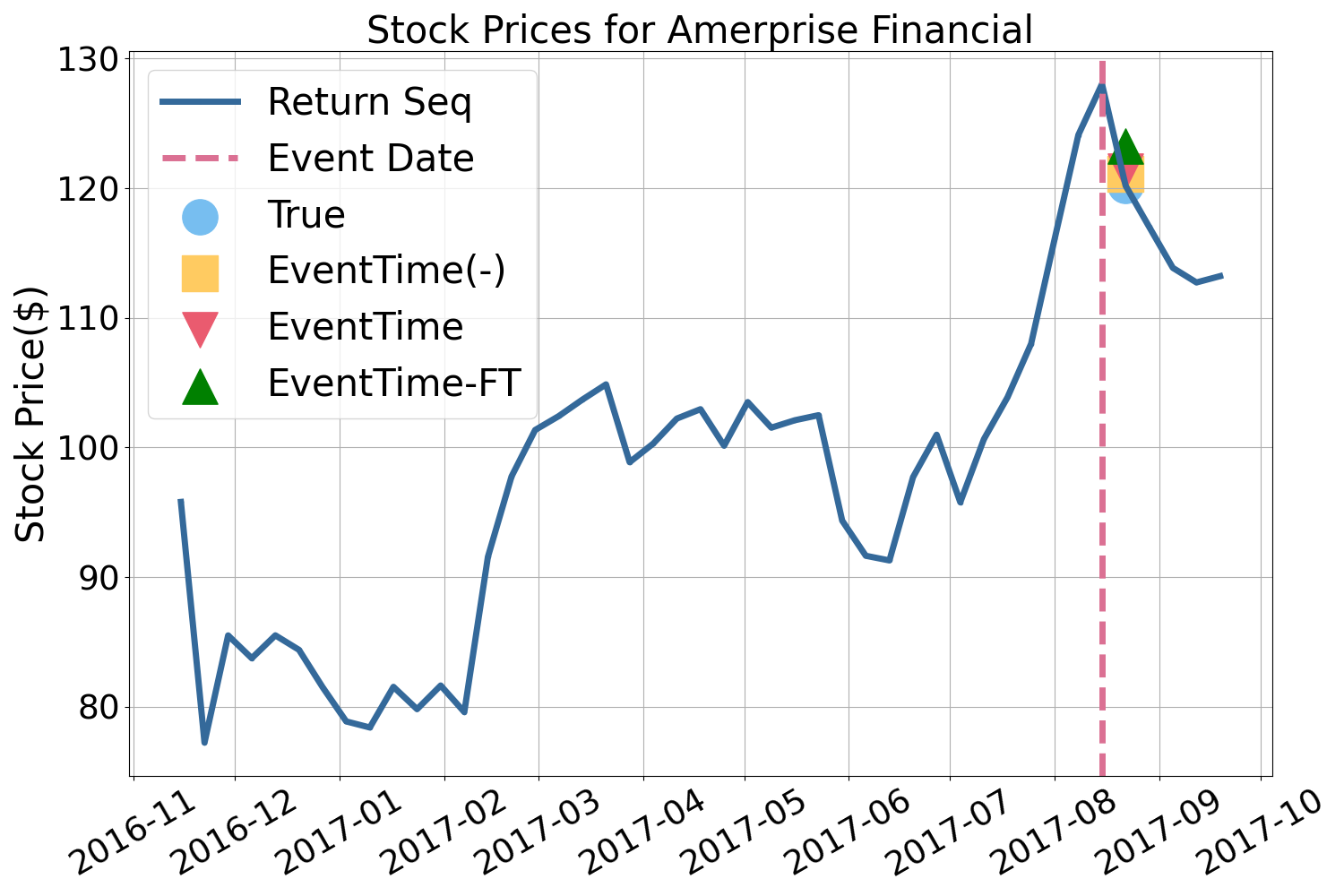}
    \includegraphics[width=0.32\textwidth]{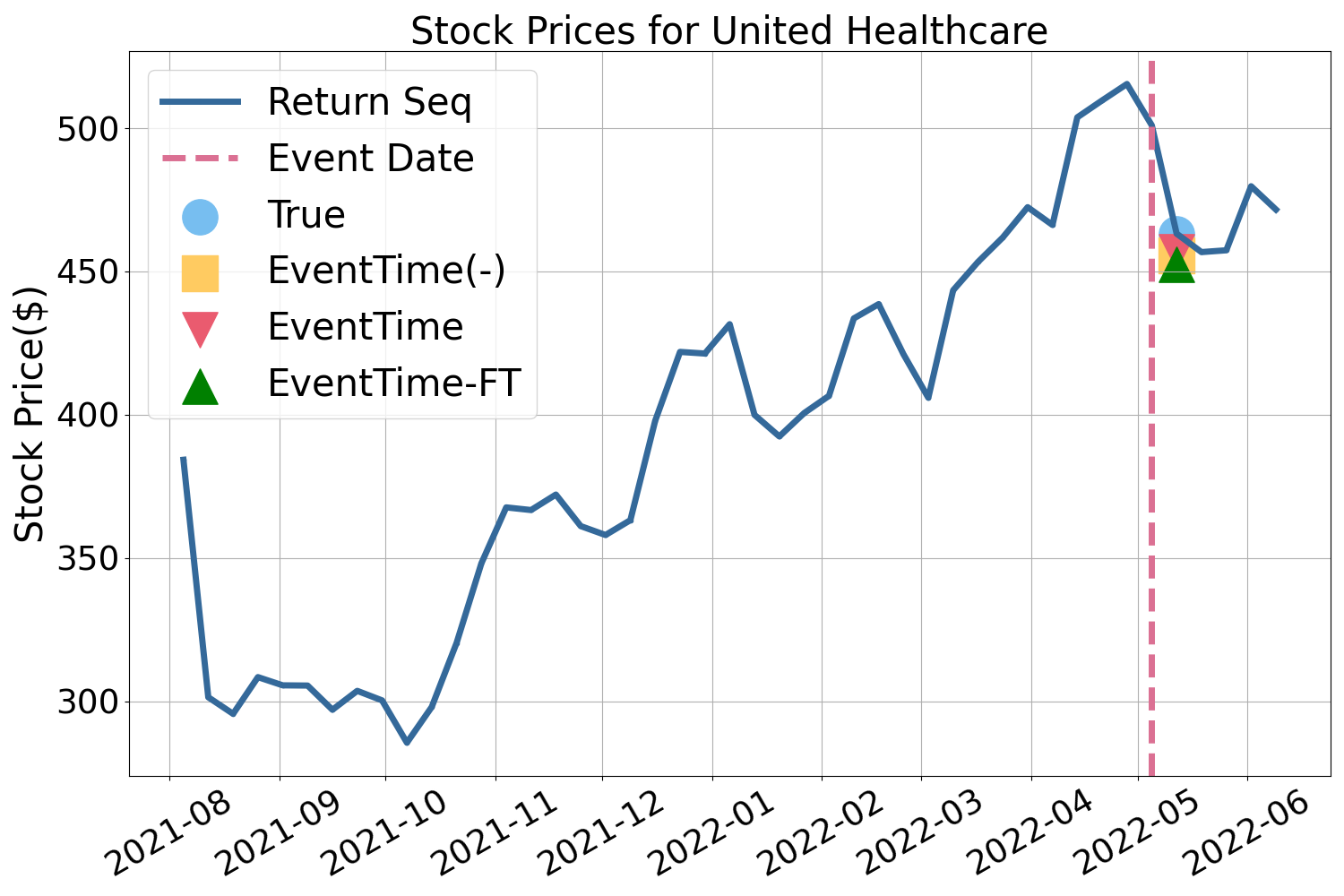}
    \caption{Representative stock-price trajectories and predicted short-term losses around cybersecurity disclosures.}
    \label{fig:case1}
\end{figure*}

\subsubsection{Effectiveness of Event-Conditioned Training (RQ2)}

Table~\ref{tab:training} summarizes the impact of progressively enhancing EventTime’s training strategy. 
Introducing event embeddings (from EventTime w/o \(e\) to EventTime(s)) improves performance, but test gains remain limited despite a substantial reduction in training loss, suggesting that naive fine-tuning does not generalize well. 
Incorporating contrastive learning (EventTime(-)) leads to a clear performance improvement, indicating that contrastive objectives help the model learn more robust event-conditioned representations, even when positive and negative pairs are defined using heuristic similarity. 
The full EventTime further integrates fused similarity by jointly considering event semantics and time-series context to construct more informative and dynamic contrastive pairs. 
This design yields additional gains in predictive accuracy and correlation metrics, highlighting the benefit of aligning multiple similarity views in event-driven learning. 
Overall, these results show that contrastive training not only strengthens the model’s use of event metadata but also guides it toward more interpretable and generalizable event-aware representations.

\subsubsection{Visualization of Event-Driven Financial Impact Prediction}

We further visualize representative post-event stock-price trajectories in Figure~\ref{fig:case1}. 
These trajectories are not used as full-sequence forecasting targets; instead, they provide qualitative context for interpreting the scalar impact estimates produced by different models. 
Across these cases, contrastive variants of EventTime produce more accurate estimates of post-event abnormal return magnitude than the supervised-only EventTime(s) baseline. 
In particular, the full EventTime model better matches the direction and scale of realized short-term losses, indicating stronger sensitivity to event-specific context. 
These qualitative results support the quantitative findings and illustrate how contrastive learning helps EventTime estimate abnormal financial impact after cybersecurity disclosures.

\subsection{Ablation Study and Analysis}

\subsubsection{Impact of Attention Depth in Event Fusion Module (RQ3)}

To identify an appropriate model complexity for event--temporal alignment, we study the effect of attention depth by varying the number of cross-attention layers from one to five, as shown in Figure~\ref{fig:attention_depth}. 
We observe a clear performance improvement when increasing the depth from one to two layers. 
A single layer provides only coarse alignment between event metadata and temporal representations, whereas the two-layer design separates event-to-temporal translation from event-guided temporal localization. 
Specifically, the first layer maps event attributes into the temporal representation space, and the second layer identifies the pre-event positions most relevant under the event context. 
In contrast, increasing the depth beyond two layers yields no further performance gains and introduces mild instability. 
This saturation suggests that two layers are sufficient to capture the event--temporal dependencies required for scalar impact prediction. 
Additional layers add unnecessary model complexity and may exacerbate overfitting under sparse event supervision. 
Accordingly, we adopt a two-layer configuration as the best trade-off between expressiveness and generalization.

\subsubsection{Feature Sensitivity Analysis of Fused Embeddings}

To understand how different training strategies affect event representation, we conduct a feature-wise perturbation analysis in Figure~\ref{fig:sensitivity}. 
For each event feature, we randomly shuffle its values across samples while keeping the remaining inputs unchanged, and then measure the resulting change in the fused embedding by its \(L_2\) norm. 
A larger change indicates that the learned representation is more sensitive to that feature.

\textbf{How does contrastive learning enhance the quality of learned event embeddings (RQ2)?}
Across all features, embeddings from the contrastively trained model, EventTime, show larger changes in response to perturbations than those from the supervised-only variant, EventTime(s). 
This indicates that contrastive learning makes the fused representation more responsive to variations in event metadata. 
In contrast, EventTime(s) exhibits uniformly lower sensitivity, suggesting that supervised fine-tuning alone may underuse event semantics under sparse supervision. 
These findings confirm that contrastive objectives encourage richer and more structured event-conditioned representations, where the model learns to distinguish meaningful event attributes from background temporal variation.

\textbf{Which event features influence the learned impact representation most (RQ3)?}
Among the event metadata, incident severity, a high-level semantic attribute derived from LLM-assisted annotation, has the strongest influence, leading to the largest change in the fused embeddings when perturbed. 
Other influential features include data sensitivity, which characterizes the type of compromised information, and affected level, which reflects the scope of impact. 
These results indicate that EventTime not only incorporates event metadata, but also assigns greater representational influence to attributes with clearer real-world relevance to financial loss. 
As a result, the learned embedding space captures a meaningful event structure in which semantically important features have stronger effects on downstream impact prediction.

\subsubsection{Missing Feature Robustness (RQ4)}

To assess robustness under partial event information, we perform an ablation study by randomly masking event metadata features. 
We vary the number of missing attributes from zero to three and evaluate the resulting performance degradation. 
As shown in Figure~\ref{fig:missing}, both EventTime and EventTime(s) exhibit declining performance as more event features are removed. 
However, the contrastively trained EventTime degrades more slowly and consistently retains stronger impact prediction performance across all missing-feature settings. 
This behavior suggests that contrastive supervision encourages more structured event-conditioned representations, enabling the model to better generalize when event descriptors are incomplete or partially corrupted.

Overall, the analysis shows that contrastive training improves EventTime in two complementary ways. 
First, it makes the fused representation more responsive to meaningful variations in event metadata. 
Second, it improves robustness when some event attributes are unavailable. 
Together, these findings support the role of dynamic contrastive learning in producing generalizable and interpretable event-aware representations for post-event impact prediction.

\subsection{Cross-Domain Generalization on Hydrological Event Impact Prediction}
\label{sec:camels}

To evaluate whether EventTime generalizes beyond cyber-financial impact prediction, we further conduct experiments on the CAMELS (Catchment Attributes and Meteorology for Large-sample Studies) dataset~\cite{newman2015development, addor2017camels}, a widely used benchmark in hydrological modeling. 
CAMELS provides long-term meteorological and hydrological records for 671 river basins across the continental United States. 
These basins span diverse geological and eco-climatological regimes, making the dataset suitable for evaluating whether an event-conditioned model can capture system responses outside the financial domain.

In this setting, we formulate a hydrological event impact prediction task. 
A precipitation spike is treated as an external event, defined as a day where observed precipitation exceeds 50 mm/day. 
Such events often trigger substantial changes in streamflow, but the magnitude of the response depends on both the antecedent hydrological state and event-day meteorological conditions. 
This provides a natural cross-domain testbed for EventTime: as in the financial setting, the model must combine historical temporal context with event-specific attributes to estimate a post-event response.

For each identified event, the model takes as input a 730-day historical window of daily streamflow measurements preceding the event day, together with a structured event metadata vector derived from event-day meteorological features, including precipitation, solar radiation, maximum and minimum temperature, and vapor pressure. 
The output is the streamflow value on the day following the event, which represents the immediate hydrological response to the precipitation shock. 
This task remains a scalar event-impact prediction problem rather than full-trajectory forecasting, and therefore provides a consistent evaluation of EventTime's event-conditioned modeling principle.

From a selected subset of 100 representative basins, we identify 1,818 extreme precipitation events spanning the period from 1 October 1999 through 30 September 2008. 
To ensure temporal generalization, we split the events chronologically: the earliest 80\% (1,416 events) are used for training, while the most recent 20\% (402 events) are reserved for testing. 
This setup reflects realistic deployment conditions, where models trained on historical events are evaluated on future unseen events.

Table~\ref{tab:camels} compares multiple backbones under the same event-conditioned impact prediction setting, as well as different training strategies applied to the EventTime architecture. 
Consistent with the results on SECURE, EventTime(s) outperforms all baseline backbones, including Transformer-based models such as Autoformer, PatchTST, iTransformer, and Pyraformer. 
This suggests that EventTime's multi-resolution encoding and event-aware fusion remain effective beyond financial time series. 
Moreover, the full EventTime model equipped with dynamic contrastive learning achieves the best performance, improving both error-based metrics and explanatory power. 
These results demonstrate that EventTime's architecture and training strategy can generalize to another domain where discrete external events perturb continuous temporal processes.

\begin{table}
    \caption{Cross-domain evaluation on CAMELS for hydrological event impact prediction.}
    \label{tab:camels}
    \centering
    \resizebox{0.33\textwidth}{!}{
    \begin{tabular}{cccc}
        \toprule
        Model & $R^2$ & MAE & RMSE\\ \midrule
        DLinear & 0.2736 & 10.9538 & 18.1245\\ \midrule
        TimeMixer & 0.2982 & 10.2326 & 17.8543\\ 
        WPMixer & 0.2822 & 10.4819 & 18.0169\\ \midrule
        Autoformer & 0.2941 & 9.5585 & 17.8676\\ 
        PatchTST & 0.3070 & 10.1851 & 17.7033\\ 
        iTransformer & 0.3078 & 10.5662 & 17.6930\\ 
        Pyraformer & 0.3192 & 9.7099 & 17.5461\\ 
        TimeXer & 0.3070 & 10.1851 & 17.7033\\ \midrule
        Text2Timeseries & 0.3325 & 9.6241 & 17.3184\\ \midrule
        EventTime(s) & 0.3570 & 9.4665 & 17.0519\\ \midrule
        EventTime(-) & 0.3836 & 8.9196 & 16.7959\\ 
        EventTime & 0.3922 & 8.7924 & 16.5795\\ 
        \bottomrule
    \end{tabular}}
\end{table}
\section{Related Work}
 
\subsection{Financial Time-series Modeling}

Financial time series are highly noisy and non-stationary, reflecting a mix of long-term macro trends, sector-level conditions, and short-term fluctuations. 
Traditional models such as ARIMA~\cite{shumway2017arima} and GARCH~\cite{bauwens2006multivariate} capture autoregressive or volatility patterns but rely on stationarity and linearity assumptions, limiting their expressiveness. 
Deep learning methods, including recurrent networks~\cite{hochreiter1997long}, temporal convolutions~\cite{bai2018tcn}, and Transformers~\cite{vaswani2017attention}, have improved nonlinear temporal modeling. 
Recent architectures such as Autoformer~\cite{wu2021autoformer}, FEDformer~\cite{zhou2022fedformer}, PatchTST~\cite{nie2023patchtst}, iTransformer~\cite{liu2024itransformer}, and TimesNet~\cite{wu2023timesnet} further emphasize decomposition, patching, or multi-resolution representation to better capture long-term and cross-scale patterns. 
However, these models mainly extrapolate endogenous dynamics from historical observations and are not designed to estimate the short-term abnormal impact of discrete external events, such as cybersecurity disclosures, regulatory actions, or unexpected firm-specific shocks.

\subsection{Event-driven Financial Modeling}

Financial markets are shaped not only by internal dynamics but also by external events. 
Early event-driven modeling showed that extracting structured events from text improves prediction of abnormal returns, volatility shifts, and risk spillovers~\cite{ding2015event,ding2016kdmevent}, while pre-trained models such as FinBERT~\cite{araci2019finbert} enhance event extraction and sentiment analysis from financial text. 
Empirical studies further show that firm-specific disclosures can generate short-term abnormal returns, such as post-earnings announcement drift~\cite{lan2024pead}, and that cybersecurity incidents can cause immediate losses followed by gradual recovery~\cite{ssrn2025cyber}. 
However, existing methods often either extract textual event features without explicitly conditioning them on the surrounding temporal state, or treat events as static auxiliary covariates, as in general exogenous-input forecasting models such as TFT~\cite{lim2021tft}. 
This overlooks the fact that the same event can induce different financial impacts depending on pre-event market context. 
EventTime addresses this gap by aligning event metadata with multi-resolution pre-event dynamics and using contrastive learning to strengthen rare-event representations.

\section{Conclusion}


In this work, we introduced \textbf{EventTime}, a framework for quantifying the financial impact of cybersecurity disclosures through event-conditioned time-series modeling. 
By combining long-term market context, short-term pre-event dynamics, and a dedicated event-fusion module, EventTime estimates how discrete external events relate to short-term abnormal financial losses. 
To address sparse event supervision and improve robustness, we incorporated dynamic contrastive learning with fused similarity, which strengthens event-aware representations by aligning similar event--market contexts. 
Experiments on real-world firm--event data show that EventTime consistently outperforms strong time-series and event-aware baselines, while case studies and ablation analyses confirm its ability to produce more accurate and interpretable estimates of post-event impact. 
These results highlight the importance of explicitly modeling exogenous events in financial time-series analysis. 
Future directions include extending EventTime to other event-impact settings, such as macroeconomic crises, policy changes, and environmental disruptions, and leveraging large language models for richer event semantics under scarce-event regimes.

\bibliographystyle{ACM-Reference-Format}
\bibliography{main}

\appendix

\newpage

\end{document}